\documentclass[letterpaper]{article} % DO NOT CHANGE THIS
\usepackage{aaai2026}  % DO NOT CHANGE THIS
\usepackage{times}  % DO NOT CHANGE THIS
\usepackage{helvet}  % DO NOT CHANGE THIS
\usepackage{courier}  % DO NOT CHANGE THIS
\usepackage[hyphens]{url}  % DO NOT CHANGE THIS
\usepackage{graphicx} % DO NOT CHANGE THIS
\usepackage{natbib}  % DO NOT CHANGE THIS AND DO NOT ADD ANY OPTIONS TO IT
\usepackage{caption} % DO NOT CHANGE THIS AND DO NOT ADD ANY OPTIONS TO IT
\usepackage{algorithm}
\usepackage{algorithmic}

\usepackage{multirow}
\usepackage{booktabs}
\usepackage{adjustbox}

\usepackage{subcaption} % 子图环境
\usepackage{newfloat}
\usepackage{listings}
\DeclareCaptionStyle{ruled}{labelfont=normalfont,labelsep=colon,strut=off} % DO NOT CHANGE THIS
\floatstyle{ruled}
\newfloat{listing}{tb}{lst}{}
\floatname{listing}{Listing}
\title{Implicit Federated In-context Learning For Task-Specific LLM Fine-Tuning}
\author{
    Dongcheng Li\textsuperscript{\rm 1},
    Junhan Chen\textsuperscript{\rm 1},
    Aoxiang Zhou\textsuperscript{\rm 1},
    Chunpei Li\textsuperscript{\rm 1},
    Youquan Xian\textsuperscript{\rm 2},
    Peng Liu\textsuperscript{\rm 1},
    Xianxian LI\textsuperscript{\rm 1}
}
\affiliations {
    \textsuperscript{\rm 1}Key Lab of Education Blockchain and Intelligent Technology, Ministry of Education, Guangxi Normal University, Guilin, Guangxi 541004, China\\
    \textsuperscript{\rm 2}School of Cyberspace Security, Beijing University of Posts and Telecommunications, Beijing 100876, China\\
}

\usepackage{bibentry}
\begin{document}

\maketitle

\begin{abstract}
As large language models continue to develop and expand, the extensive public data they rely on faces the risk of depletion. Consequently, leveraging private data within organizations to enhance the performance of large models has emerged as a key challenge. The federated learning paradigm, combined with model fine-tuning techniques, effectively reduces the number of trainable parameters. However,the necessity to process high-dimensional feature spaces results in substantial overall computational overhead. To address this issue, we propose the Implicit Federated In-Context Learning (IFed-ICL) framework. IFed-ICL draws inspiration from federated learning to establish a novel distributed collaborative paradigm, by converting client local context examples into implicit vector representations, it enables distributed collaborative computation during the inference phase and injects model residual streams to enhance model performance. Experiments demonstrate that our proposed method achieves outstanding performance across multiple text classification tasks. Compared to traditional methods, IFed-ICL avoids the extensive parameter updates required by conventional fine-tuning methods while reducing data transmission and local computation at the client level in federated learning. This enables efficient distributed context learning using local private-domain data, significantly improving model performance on specific tasks.
\end{abstract}


\section{Introduction}

In recent years, the rapid development of Large Language Models (LLMs) has brought about a revolutionary transformation in the field of artificial intelligence. These models have not only pushed the boundaries of technology but also reshaped the fundamental paradigm of human-computer interaction. Trained on massive textual datasets and built upon deep neural network architectures, models such as the GPT series, LLaMA, and PaLM have demonstrated unprecedented capabilities in language understanding and generation, surpassing the limitations of traditional natural language processing approaches. As the number of parameters has scaled from billions to hundreds of billions, LLMs have exhibited remarkable emergent abilities, such as reasoning, planning, coding, and cross-modal understanding. The widespread adoption of LLMs has permeated various domains, including intelligent dialogue systems, creative writing assistance, medical diagnosis support, scientific research acceleration, and personalized education. These applications not only enhance efficiency and innovation but also provide new tools for addressing the complex challenges faced by humanity.

As the parameters and functionalities of LLMs continue to grow,their rate of data consumption is also increasing rapidly. Studies have shown that publicly available high-quality textual data is expected to be exhausted between 2026 and 2032 \cite{ye2024openfedllmtraininglargelanguage}. Reusing existing datasets not only limits the upper bound of model performance but also risks overfitting and reduces the model’s generalization ability. According to a survey conducted by Epoch AI, the total global volume of textual data is approximately 31 trillion tokens, with publicly available data comprising only a small fraction. In contrast, private data, which is often of higher quality and more domain-specific, has emerged as the new frontier for breakthroughs in model performance \cite{jones2024ai}. As a result, Federated Learning(FL) has emerged as a highly attractive solution, enabling users to supplement large models with knowledge derived from privately held data through collaborative multi-party training. This approach effectively enhances the reasoning capabilities of LLMs for specific tasks.

To improve the performance of foundation models on specific tasks, two main paradigms have emerged: fine-tuning and In-Context Learning (ICL) \cite{brown2020language}. Fine-tuning approaches include full fine-tuning and parameter-efficient fine-tuning methods such as LoRA\cite{hu2022lora} and P-Tuning-v2\cite{liu2021p}. In contrast, In-Context Learning is a training-free approach that guides the model to perform specific tasks by providing a few examples during inference, significantly lowering the barrier to entry for task adaptation.

To effectively leverage high-quality private data within organizations, recent studies \cite{peng2024fedpft,wu2024fedbiot} have proposed combining FL with fine-tuning techniques. While federated fine-tuning can adapt LLM to downstream tasks, it requires updating model parameters within high-dimensional feature spaces, leading to significant computational overhead. Consequently, the deployment of Federated Large Language Models (FedLLMs) faces substantial challenges in practice. One significant barrier is the communication cost. For instance, transmitting the parameters of LLaMA3.1-405B over a 100 Mbps network would require more than 36 hours, which far exceeds the capacity of contemporary communication systems. On the other hand, approaches based on ICL \cite{wu2024federated,DBLP:journals/corr/abs-2403-06131} involve collecting examples from multiple clients and incorporating them into the inference prompt. However, this fundamentally violates the principle of data locality in federated learning and poses a serious risk of sensitive information leakage.

To address the limitations of ICL, \cite{li2024implicit} offers a novel perspective by converting contextual examples into vector representations and injecting them into LLMs to perform inference tasks. This work provides valuable inspiration for our research. Building upon it, we propose a collaborative framework named Implicit Federated In-Context Learning, which aims to tackle the dual challenges of computational inefficiency in traditional federated fine-tuning and the limited collaborative capacity of conventional ICL. Unlike traditional FL paradigms that require extensive local training on client devices, our approach innovatively decomposes the federated process into two parts: the extraction of context vectors and the collaborative computation of injection coefficients. This significantly reduces the computational burden on local clients.

We implement efficient collaboration through a three-stage process. In the first stage, each participating client designs task-specific context templates, converts local data into context vectors, and sends them to the server for aggregation. In the second stage, the server returns the aggregated global context vectors to each client, where clients compute the perplexity loss using their local data to calibrate the injection coefficients. These coefficients are then sent back to the server for aggregation. After several rounds of iterative optimization, the injection coefficients are refined to ensure optimal integration of contextual information into the LLM. Finally, in the third stage, the server distributes the calibrated coefficients to clients, which can then convert the raw LLM into a task-specific LLM with a single linear injection operation. This design not only reduces computational and communication overhead but also achieves an effective decoupling of data utilization and model training, offering a new paradigm for distributed AI collaboration in resource-constrained environments.The main contributions of this paper are as follows:

\begin{itemize}
    \item we propose a novel federated ICL paradigm. Instead of synthesizing contextual data, our method transmits and aggregates context vectors, which are then injected during the model inference phase to enhance performance.
    
    \item In contrast to traditional FL, our approach decomposes the federated process into two components: context vector aggregation and injection coefficient training. Clients are responsible for converting local data into context vectors and performing lightweight training of injection coefficients. This design significantly reduces communication bandwidth requirements and computational overhead, enabling effective participation from resource-constrained devices.
    
    \item Extensive experiments across multiple text classification tasks demonstrate that, compared to federated parameter fine-tuning, IFed-ICL reduces clients computational overhead by more than 20 times and communication costs by approximately $10^4$ times, thereby ensuring the feasibility of large-scale federated deployment.
\end{itemize}

\section{Related Work}
\subsection{Federated Fine-Tuning}

FL is a distributed machine learning paradigm that enables multiple clients to collaboratively train a global model by exchanging model parameters without exposing their raw local data\cite{mcmahan2017communication,konevcny2016federated,yang2019federated}. Its primary goal is to mitigate the systemic privacy risks inherent in traditional centralized data collection \cite{kairouz2021advances}. In recent years, LLMs have achieved remarkable breakthroughs in performance. However, the scale of publicly available datasets has approached its limit, and further development of these models is increasingly constrained by the challenge of "data silos" \cite{villalobos2022will}. Federated Large Language Models (FedLLMs) \cite{chen2023federated} have been proposed in this context, aiming to combine the powerful generalization capabilities of LLMs with the privacy-preserving advantages of FL.

Nevertheless, the implementation of FedLLMs continues to face formidable challenges. When the number of model parameters reaches the scale of billions, full-parameter fine-tuning of LLMs results in massive communication overhead, which severely limits their scalability in practical deployments \cite{shu2024ferret}. To address this, Parameter-Efficient Fine-tuning (PEFT) has become the mainstream optimization pattern \cite{hu2024federated}. The core idea is to freeze the majority of the LLM’s parameters and fine-tune only a small set of newly added or selectively chosen parameters, thereby reducing the computational, storage, and communication burden on client devices. The FedPETuning framework \cite{zhang2023fedpetuning} was among the earliest to systematically incorporate multiple PEFT methods (including LoRA) into FL settings, demonstrating the feasibility of significantly lowering communication costs by aggregating only a small portion of trainable parameters.

Beyond LoRA and its variants, other PEFT strategies have also provided diverse and efficient fine-tuning paths for FedLLMs. In adapter-based methods, FedAdapter \cite{cai2022fedadapter} enhances training efficiency by dynamically configuring adapters and leveraging activation caching, while FeDeRA \cite{yan2024federa} initializes low-rank adapters via singular value decomposition (SVD) to improve performance under Non-IID data distributions. Prompt-based methods have gained significant attention due to their extremely low communication overhead. For example, FedPepTAO \cite{che2023federated} achieves efficient fine-tuning through partial prompt tuning and dual-end adaptive optimization. Additionally, some selective PEFT methods, such as BitFit \cite{zaken2021bitfit}, which only fine-tunes bias parameters, and FedAMoLE \cite{zhang2024personalized}, which builds mixtures of LoRA experts for highly heterogeneous scenarios, demonstrate strong potential in improving both personalization and efficiency.

Furthermore, Federated Knowledge Distillation (FKD) offers an alternative approach to reducing communication overhead by allowing clients to transmit compact representations of knowledge instead of full model parameters. For instance, AdaFedSelecKD \cite{feng2024adapter} adopts adapter-based selective knowledge distillation to improve communication efficiency. These communication optimization techniques are not mutually exclusive with PEFT methods.Despite a series of advancements in efficient parameter tuning, the process fundamentally relies on executing gradient updates and backpropagation within high-dimensional parameter spaces. Consequently, the reduction in computational overhead is inherently constrained by the upper limit of the trainable rank.

\subsection{In-Context Learning}
Compared to parameter fine-tuning,which requires explicitly updating model weights via backpropagation,ICL enables LLMs to adapt to new tasks during inference simply by providing demonstration examples, without modifying any model parameters. This opens up a new lightweight deployment pathway for LLMs. The study by \cite{von2023transformers} reveals that the Transformer architecture can implicitly simulate gradient descent dynamics during inference, offering key insights into the underlying mechanism of ICL. Complementarily, \cite{wies2023learnability} provides a rigorous theoretical perspective based on Probably Approximately Correct learning theory, emphasizing the decisive impact of context example quality on ICL performance and laying a foundational framework for subsequent methodological innovations.

In the domain of ICL optimization, the empirical study by \cite{agarwal2024many} demonstrates that increasing the number of in-context examples initially improves model performance on open-ended tasks. \cite{bertsch2024incontext} further investigates the model’s capacity to utilize extended context windows containing numerous examples. To address the limitation of context window length, \cite{ye2023compositional} proposes a retrieval-augmented approach that dynamically selects task-relevant exemplars, significantly enhancing inference accuracy. However, such strategies rely heavily on the precision of the retrieval system and inevitably introduce additional architectural complexity. \cite{huang2024multimodal} explores techniques for compressing multiple examples into compact latent representations, thereby reducing inference costs and improving performance in multimodal ICL settings. Nonetheless, practical deployment is constrained by the requirement to access internal model states. \cite{hendel2023context} conceptualizes the transformation of multiple in-context examples into a single task vector to guide model inference, thereby simplifying traditional explicit example enumeration. Expanding on this line of research, \cite{li2024implicit} injects task-relevant context directly into intermediate activation layers of the model. This strategy significantly reduces reliance on long contexts and improves computational efficiency.

In FL scenarios, ICL can be leveraged to enhance model personalization while avoiding the direct sharing of sensitive data. Methods such as \cite{wu2024federated,DBLP:journals/corr/abs-2403-06131} combine the context learning capabilities of LLMs with FL by having servers and clients share LLM-generated synthetic data, allowing clients to fine-tune using locally private datasets. \cite{wang2025federated} achieves efficient and low-overhead learning for question-answering tasks in distributed environments by applying ICL using high-quality local data at each client, and employing a parameter-free communication strategy. However, these approaches commonly assume an infinite context window, overlooking truncation bias and exemplar overflow effects caused by token limits, which results in significant performance degradation in long-sequence scenarios.

\section{Proposed Framework} 
\begin{figure*}[htb]
    \centering
    \includegraphics[height=0.4\textheight]{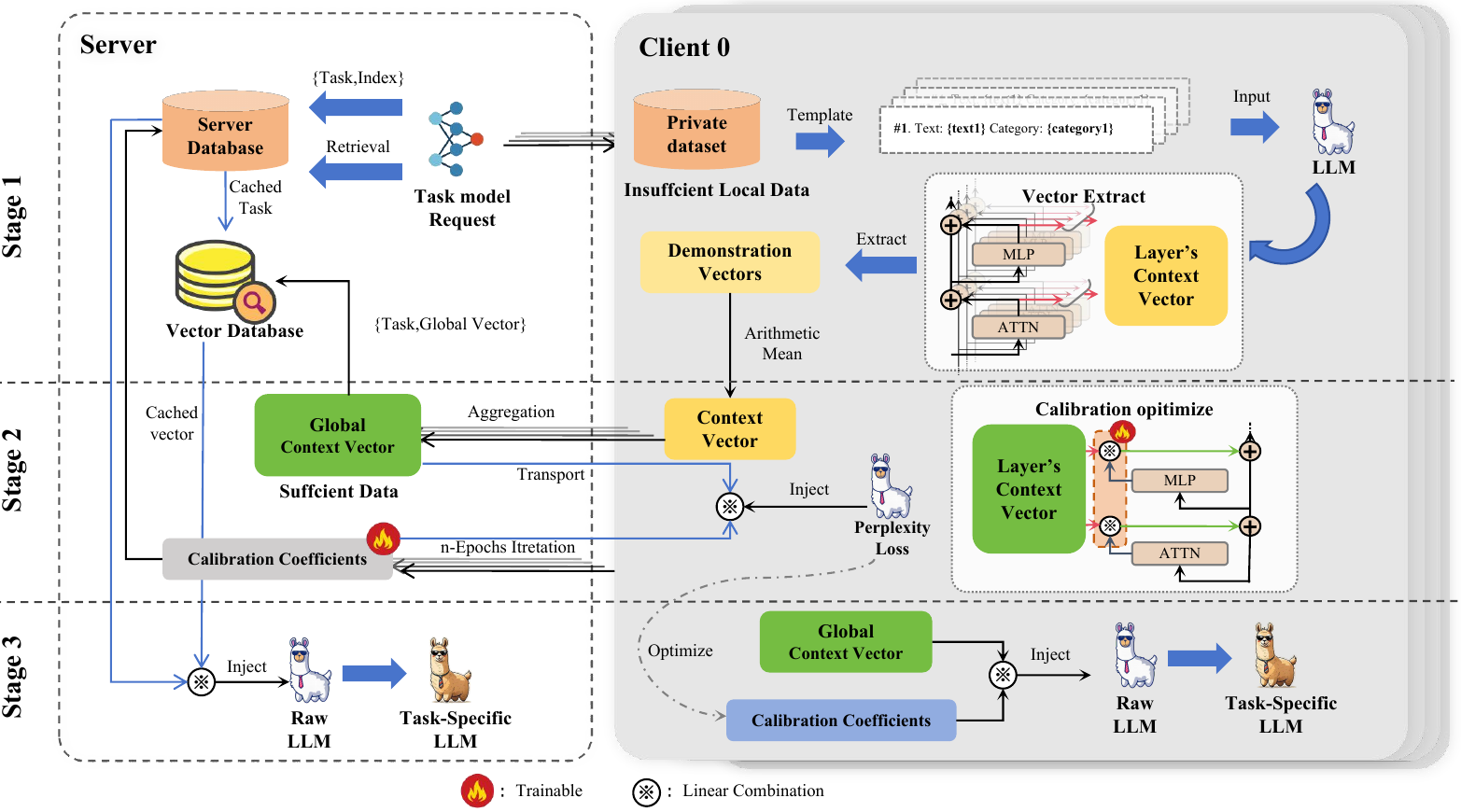}
    \caption{Overall of IFed-ICL framework.First stage: Each client extracts vector representations from its private dataset $\mathcal{D}_k$ and transmits them to the server for aggregation. Second stage: Clients collaboratively calibrate the injection coefficients $\Lambda$ by minimizing the perplexity loss on their local data in coordination with the server. Third stage: The trained injection coefficients and the aggregated global context vector are utilized to enable in-context learning for the large language model.
}
    \label{fig:framework-ifed-icl}
\end{figure*}

In this section, we present the workflow of IFed-ICL. As illustrated in Figure~\ref{fig:framework-ifed-icl}, the overall collaborative framework of IFed-ICL is divided into three key stages.
\subsection{System Setup}
In IFed-ICL, the system consists of $\mathcal{K}$ clients and a central server. The server maintains both a conventional storage database and a vector database to store task-related information and aggregated context vectors. Each client possesses its own private dataset $\mathcal{D}_k$and a pre-trained LLM $\mathcal{M}$. Based on the requirements of a specific task, the server defines a context example template $\mathcal{T}$ and distributes it to the clients. Each client then uses the template to label relevant data from its local private dataset  $\mathcal{D}_k$as context examples $E_k = \{(s_{k,j}, o_{k,j})\}_{j=1}^{N_k}$, where $(s_{k,j}, o_{k,j})$ denotes the $j$-th input-output pair and $N_k$ is the number of examples generated by client $k$. The ultimate goal is to leverage these private context examples from multiple clients to collaboratively enhance the large model’s representational capacity and generalization ability.

\subsection{Phase 1: Context Vector Extraction and Upload}
At the beginning of each round of collaboration in IFed-ICL, the server and the selected participating clients $k \in \mathcal{K}$ first perform forward propagation using the large language model $\mathcal{M}$ on each demonstration example $(s_{k,j}, o_{k,j})$. During this process, the context examples are used at the token positions required for prediction in each layer of the model to extract intermediate activation vectors. These activation vectors include the outputs from the Multi-Head Attention (MHA) and Multi-Layer Perceptron (MLP) modules across all $L$ Transformer layers. Let the MHA and MLP activations extracted from the $l$-th layer of example $s_{k,j}$ be denoted as $a_{k,j,l}^e$ and $m_{k,j,l}^e$, respectively. These layer-wise activations are then fused to form the demonstration vector for the example:
\begin{equation}
    d'_{k,j} = \{a_{k,j,l}^e, m_{k,j,l}^e\}_{l=1}^L
\end{equation}

Finally, the server and client $k$ compute the arithmetic mean of all locally generated demonstration vectors $\{d'_{k,j}\}_{j=1}^{N_k}$ to obtain the local context vector $v_k$:
\begin{equation}
    v_k = \frac{1}{N_k} \sum_{j=1}^{N_k} d'_{k,j}
    \label{eq:local_context_vector}
\end{equation}
where the averaging is performed element-wise across each dimension of the vectors. After generating $v_k$, the client uploads it to the central server. Since $v_k$ is a compact vector representation, its size is significantly reduced compared to the original data or large-scale model parameters.

\subsection{Stage 2: Global Context Vector Aggregation and Coefficient Calibration}
After receiving the local context vectors ${v_k^t}$ from all participating clients $k \in \mathcal{K}_t$, the central server first aggregates these vectors to form a global context vector $v_g^t$. The aggregation method adopts Federated Averaging:

\begin{equation}
    v_g^t = \frac{1}{|\mathcal{K}_t|} \sum_{k \in \mathcal{K}_t} v_k^t
    \label{eq:global_context_vector}
\end{equation}

Let the corresponding components of the $l$-th layer MHA and MLP of the global context vector $v_g^t$ be denoted as $(\overline{a}_{l}^e)_g^t$ and $(\overline{m}_{l}^e)_g^t$.

By configuring a set of hyperparameters $\Lambda$, the model's residual stream during inference can be injected with contextual information to achieve the goal of in-context learning.
\begin{equation}
\Lambda = \{\lambda_l^a, \beta_l^a, \lambda_l^m, \beta_l^m\}_{l=1}^L
\end{equation}
The client first initializes the hyperparameters $\Lambda_k$ and uses the local dataset $E_k$ to optimize and calibrate the injection coefficients of the context vector. For each calibration sample $(x_i^k, y_i^k) \in \mathcal{D}k$, the client feeds it into the model $\mathcal{M}$ and performs layer-wise forward propagation. Assume that for the query $x_i^k$, the MHA and MLP activations at layer $l$ and token position $\tau$ are denoted as $a_{l,\tau}^k$  and $m_{l,\tau}^k$ respectively. Then, the updated residual stream after injection $r_{l,\tau}^k$ is:
\begin{equation}
    \resizebox{0.9\columnwidth}{!}{%
        $r_{l,\tau}^k \leftarrow r_{l-1,\tau}^k + (\lambda_{l}^{a}(\overline{a}_{l}^e)_g^t + \beta_{l}^{a}a_{l,\tau}^k) + (\lambda_{l}^{m}(\overline{m}_{l}^e)_g^t + \beta_{l}^{m}m_{l,\tau}^k)$%
    }
    \label{eq:i2cl_injection_fed}
\end{equation}

Subsequently, all clients perform $n$ rounds of joint training with the server, optimizing the injection coefficients $\Lambda_k$ by minimizing the perplexity loss on the calibration dataset $\mathcal{D}_k$.

\begin{equation}
\Lambda_k^{(n)} = \arg\min_{\Lambda} \left\{
-\!\sum_{(x_i^k,y_i^k)\in\mathcal{D}_k}\!\!
\log P(y_i^k|x_i^k,v_g,\Lambda_k) \right\}
\label{eq:calibration_loss}
\end{equation}
Upon receiving the local injection coefficients from all participating clients, the server aggregates them to compute the global injection coefficients.
$\Lambda_g^{(n)}$:
\begin{equation}
    \Lambda_g^{(n+1)} = \frac{1}{K} \sum_{k=1}^{K} \Lambda_k^{(n)}
\end{equation}

After receiving the global injection coefficients returned by the server, each client uses Equation \ref{eq:calibration_loss} to iteratively compute $\Lambda_k^{(n)}$. After $n$ iterations, the training is completed. This optimization process only targets the injection coefficients $\Lambda$, whose number is significantly smaller than that of the LLM parameters. For the global context vector, only a single round of aggregation is required. Furthermore, this approach can adapt to incremental data scenarios, where only the incremental context vectors need to be aggregated over multiple rounds, and this can be computed in parallel with the calibration of injection coefficients.

\subsection{Stage 3: Global Calibration Coefficients Injection}
After calibration is completed, the central server distributes the optimized injection coefficients $\Lambda_g^*$ obtained in the current round to all participating clients. Upon receiving $\Lambda_g^*$, each client applies it to their local LLM $\mathcal{M}$. When new inference queries arrive, the client’s LLM performs context injection according to Equation~\ref{eq:i2cl_injection_fed} using $v_g$ and $\Lambda_g^*$. Through this single linear injection operation, the raw LLM is transformed into a task-specific LLM tailored for specific scenarios, without requiring any local parameter fine-tuning or gradient computations. This design effectively decouples data utilization from model training, significantly reducing computational and communication overhead on the client side, while also addressing the issue of token length limitations in long context inputs. It thus provides a novel paradigm for distributed AI collaboration in resource-constrained environments.

\textbf{Application Optimization:} We designed two types of databases to enhance task processing efficiency and resource utilization. The first type is a server database, such as MongoDB, which is utilized to store task-related information along with their corresponding calibration coefficients. The second type is a vector database, such as Elasticsearch, designed to store global context vectors. By establishing a task vector index, the similarity search process is significantly reduced compared to traditional Retrieval-Augmented Generation (RAG) systems.Specifically,upon receiving a new task request, the server first queries the task index to retrieve the associated task information and calibration coefficients. If the task has been previously cached, the server directly retrieves the calibrated coefficients and the corresponding context vector from the databases. Otherwise, the server collects sufficient data and performs an $n$-epoch iterative training process to obtain a task-specific LLM. During this process, the server also updates the calibration coefficients and the global context vector, storing the newly generated data into the respective databases for efficient future retrieval. This caching mechanism effectively reduces redundant computation and significantly accelerates task response time, forming a closed-loop optimization system.

\section{Experiments}
\subsection{Experimental Setup}
To comprehensively evaluate the effectiveness of our proposed IFed-ICL framework, we conduct rigorous experiments using two LLMs, LLaMA-3-8B and Qwen2.5-7B on three widely used text classification datasets: SUBJ \cite{pang2004sentimental}, Emotion \cite{chatterjee2019semeval}, and AG News \cite{chatterjee2019semeval}. Except for the zero-shot baseline, which is evaluated using only 500 test samples to assess the model's inherent capabilities, all other experiments utilize 5,000 training samples and 500 test samples from each dataset.

To simulate the commonly observed non-independent and identically distributed (Non-IID) nature of federated learning, we partition the training data across 10 clients using a Dirichlet distribution with a concentration parameter of $\alpha = 0.5$. We compare IFed-ICL against several representative baselines:

\textbf{Zero-Shot}: serving as a reference for the model's raw performance;
 
\textbf{Local ICL} \cite{brown2020language}: which simulates a non-collaborative scenario where each client performs inference independently using local data;

\textbf{FedAvg-LoRA} \cite{hu2022lora}: a representative method of parameter-efficient fine-tuning (PEFT) in federated settings, where clients fine-tune LoRA weights that are aggregated via federated averaging on the server.

Zero-shot performance serves as a baseline for evaluating the inherent task capabilities of LLMs without any task-specific adaptation. To highlight the value of Federated ICL, we design comparative experiments built upon this baseline. In contrast to the traditional paradigm of federated learning in PEFT, which relies on exchanging model parameters, we introduce three comparative baselines to demonstrate the superiority of our proposed framework.All server-side computations are executed on a single NVIDIA A100 GPU.

\begin{table*}[htbp]
\centering
\caption{Communication overhead of FedAvg-LoRA and IFed-ICL on the SUBJ dataset.}
\begin{tabular}{ccll}
\toprule
\textbf{Method} & \textbf{Direction} & \textbf{FedAvg-LoRA} & \textbf{IFed-ICL} \\
\midrule
\multirow{2}{*}{\textbf{Initialization}} 
    & Client $\rightarrow$ Server & 0 KB& 514 KB (context vector) \\
\cmidrule(l){2-4}
    & Server $\rightarrow$ Client & 13357.78 KB & \begin{tabular}[c]{@{}l@{}} \textbf{515.8 KB} (context vector + calibration coefficients) \end{tabular} \\
\midrule
\textbf{Training (per round)} 
    & Client $\leftrightarrow$ Server & 13357.78 KB& \textbf{1.8 KB} (calibration coefficients ) \\
\bottomrule
\end{tabular}
\label{tab:communication_breakdown_detailed}
\end{table*}
\subsection{Experimental Design}
IFed-ICL is designed to address key challenges in federated large language models  related to performance, efficiency, and collaborative adaptation. We evaluate its effectiveness through the following aspects:

\textbf{Performance Comparison:} We compare the task accuracy of IFed-ICL against several baselines, including Zero-Shot, Local ICL, and a representative parameter-efficient federated fine-tuning method, FedAvg-LoRA. The core objective is to assess whether our training-free federated paradigm can achieve competitive or superior performance relative to computationally intensive fine-tuning approaches.

\textbf{System Efficiency Evaluation:} This aspect focuses on the practical deployability of the framework. We quantitatively compare the communication overhead (in KB per round) and the total client-side computation time (in seconds) per federated round, demonstrating IFed-ICL’s suitability for deployment in resource-constrained environments.

\textbf{Impact of Federated Aggregation on Injection Coefficient Performance: }This analysis evaluates the effectiveness of federated aggregation in producing a global injection coefficient from clients’ locally calibrated coefficients. Specifically, we compare the performance of the global coefficient obtained via aggregation with that of locally optimized coefficients used independently by each client. The goal is to quantitatively demonstrate that federated aggregation can effectively integrate diverse local knowledge, resulting in a superior injection coefficient that enhances both model performance and generalization.

% \textbf{Impact of Federated Rounds on Injection Coefficient Performance:} This investigation focuses on the dynamic evolution of the federated learning process. We track how the global injection coefficient improves over successive communication rounds and how clients performance changes across iterations. This analysis aims to assess the convergence and stability of IFed-ICL whether multi-round collaboration leads to continuous refinement of the injection coefficient and ultimately achieves a stable and high-performing state. Understanding this convergence behavior is critical for characterizing training dynamics and identifying the optimal number of communication rounds.

\begin{table}[ht]
\centering
\caption{Performance comparison of Llama-3-8B and Qwen2.5-7B on the SUBJ, Emotion, and AG News datasets. Accuracy (acc) and F1-score are reported in percentage (\%)}
\begin{adjustbox}{max width=\linewidth}
\begin{tabular}{llcccc}
\toprule
\multirow{2}{*}{\textbf{Dataset}} & \multirow{2}{*}{\textbf{Method}} & \multicolumn{2}{c}{Llama-3-8B} & \multicolumn{2}{c}{Qwen2.5-7B} \\
\cmidrule(lr){3-4} \cmidrule(lr){5-6}
& & acc (\%) & f1 (\%) & acc (\%) & f1 (\%) \\
\midrule
\multirow{4}{*}{SUBJ}
& Zero-Shot & 62.60 & 62.48 & 62.60 & 62.48\\
& Local ICL & 70.00 &  66.90 & 70.80 & 67.90\\
& FedAvg-LoRA & 66.00 & 64.58 & 66.00 & 64.58\\
& \textbf{IFed-ICL} & \textbf{91.20} & \textbf{90.67} & \textbf{81.20} & \textbf{80.66}\\
\midrule
\multirow{4}{*}{Emotion}
& Zero-Shot & 52.20 & 53.51 & 52.20 & 53.52\\
& Local ICL & 49.82 & 50.42 & 49.70 & 50.30\\
& FedAvg-LoRA & 54.60 & 54.89 & 54.60 & 54.89\\
& \textbf{IFed-ICL} & \textbf{67.40} & \textbf{65.85} & \textbf{60.80} & \textbf{59.32}\\
\midrule
\multirow{4}{*}{AG News}
& Zero-Shot & 82.40 & 80.06 & 82.40 & 82.04\\
& Local ICL & 75.00 & 74.70 & 74.90 & 74.70\\
& FedAvg-LoRA & 79.00 & 78.58 & 80.00 & 79.60\\
& \textbf{IFed-ICL}  & \textbf{91.60} & \textbf{89.57} & \textbf{90.60} & \textbf{90.53}\\
\bottomrule
\end{tabular}
\end{adjustbox}
\label{tab:performance_acc_f1}
\end{table}

\begin{table}[ht]
\centering
\caption{Running time (seconds) of Llama-3-8B and Qwen2.5-7B on SUBJ, Emotion, and AG News.}
\begin{adjustbox}{max width=\linewidth}

\begin{tabular}{llcc}
\toprule
\multirow{2}{*}{\textbf{Dataset}} & \multirow{2}{*}{\textbf{Method}} & \multicolumn{2}{c}{\textbf{Running time (s)}} \\
\cmidrule(lr){3-4}
& & Llama-3-8B & Qwen2.5-7B \\
\midrule
\multirow{4}{*}{SUBJ}
& Zero-Shot & 30.14 & 29.44 \\
& Local ICL & 45.74 & 45.31 \\
& FedAvg-LoRA & 27637.48 & 14945.92 \\
& \textbf{IFed-ICL} & \textbf{670.25} & \textbf{530.99} \\
\midrule
\multirow{4}{*}{Emotion}
& Zero-Shot & 28.92 & 57.74 \\
& Local ICL & 40.82 & 84.19 \\
& FedAvg-LoRA & 28168.84 & 14043.47 \\
& \textbf{IFed-ICL} & \textbf{809.73} & \textbf{1298.17} \\
\midrule
\multirow{4}{*}{AG News}
& Zero-Shot & 27.90 & 56.45 \\
& Local ICL & 54.02 & 112.20 \\
& FedAvg-LoRA & 30046.70 & 13719.66 \\
& \textbf{IFed-ICL} & \textbf{973.62} & \textbf{867.55} \\
\bottomrule
\end{tabular}
\end{adjustbox}
\label{tab:runtime_only}
\end{table}

\subsection{Performance Comparison}
As shown in Table~\ref{tab:performance_acc_f1}, our proposed IFed-ICL significantly outperforms all baseline methods across all evaluated tasks. Notably, on AG News, FedAvg-LoRA underperforms even compared to Local ICL and Zero-Shot. This can be attributed to its reliance on client-specific fine-tuning using local data. Due to data heterogeneity, such local adaptation may lead to overfitting on certain clients, thus degrading the overall performance. In contrast, ICL and Zero-Shot approaches primarily leverage global knowledge and exhibit stronger generalization to the target task, making them more robust to data heterogeneity. By aggregating and injecting context vectors, our proposed method effectively mitigates the adverse effects of non-IID data, thereby achieving superior performance.

\begin{figure*}[htbp]
\centering
%--- 第 1 张 ---
\begin{subfigure}[b]{0.32\textwidth}
  \centering
  \includegraphics[width=\textwidth, height=4.4cm, keepaspectratio]{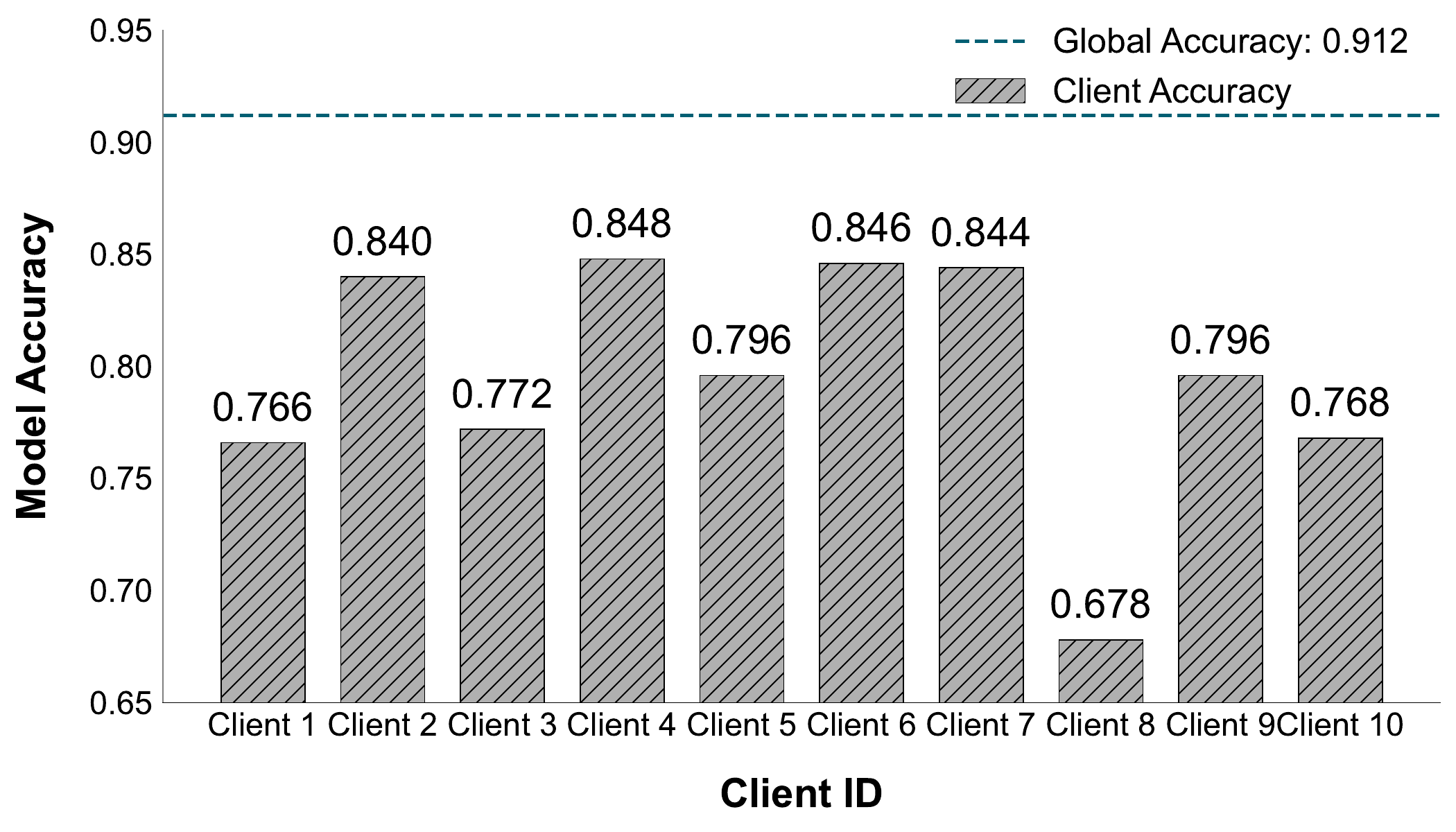}
  \caption{AG News}
\end{subfigure}
\hfill   % 让三张图均匀分布
%--- 第 2 张 ---
\begin{subfigure}[b]{0.32\textwidth}
  \centering
  \includegraphics[width=\textwidth, height=4.4cm, keepaspectratio]{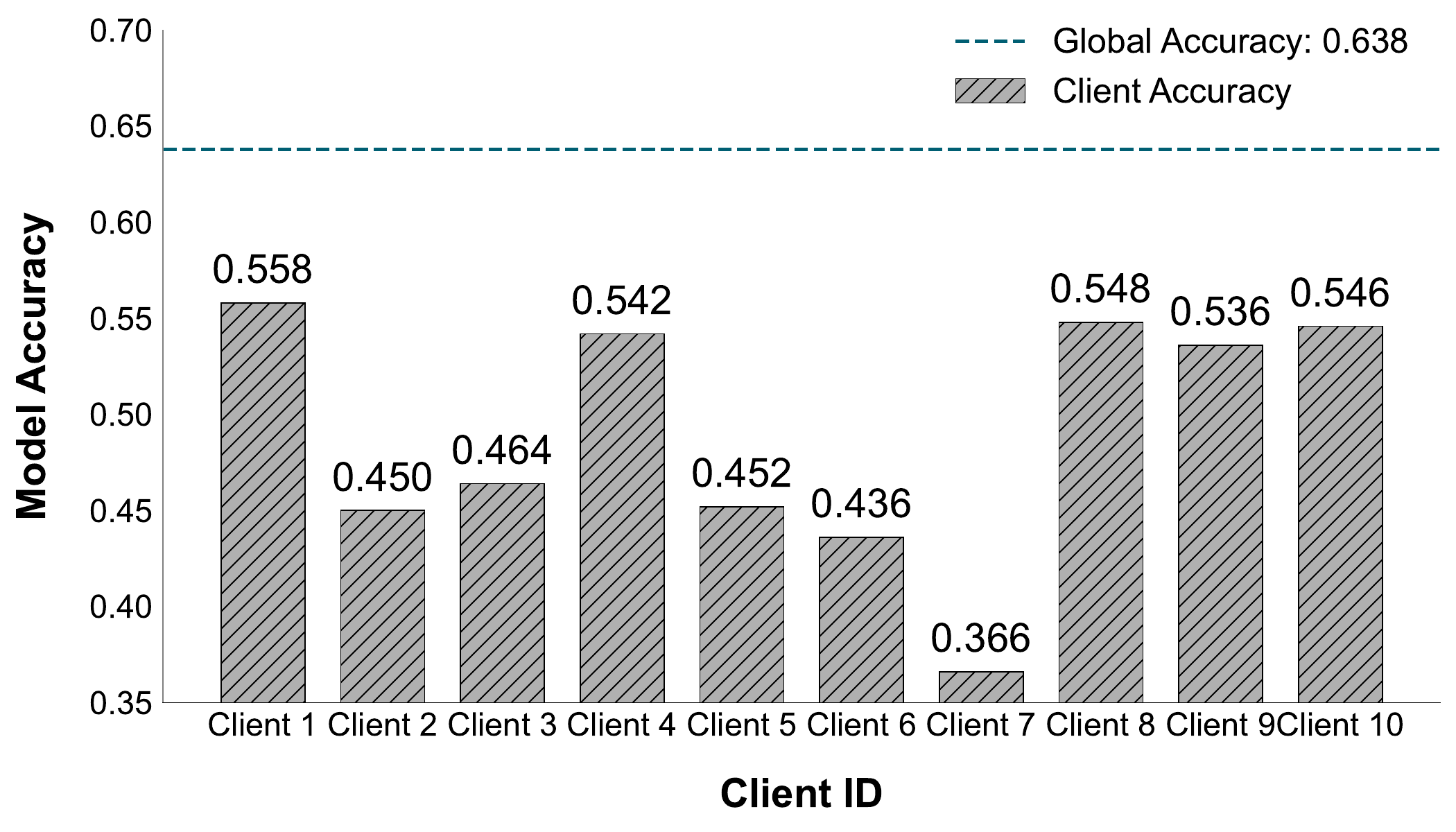}
  \caption{Emotion}
\end{subfigure}
\hfill
%--- 第 3 张 ---
\begin{subfigure}[b]{0.32\textwidth}
  \centering
  \includegraphics[width=\textwidth, height=4.4cm, keepaspectratio]{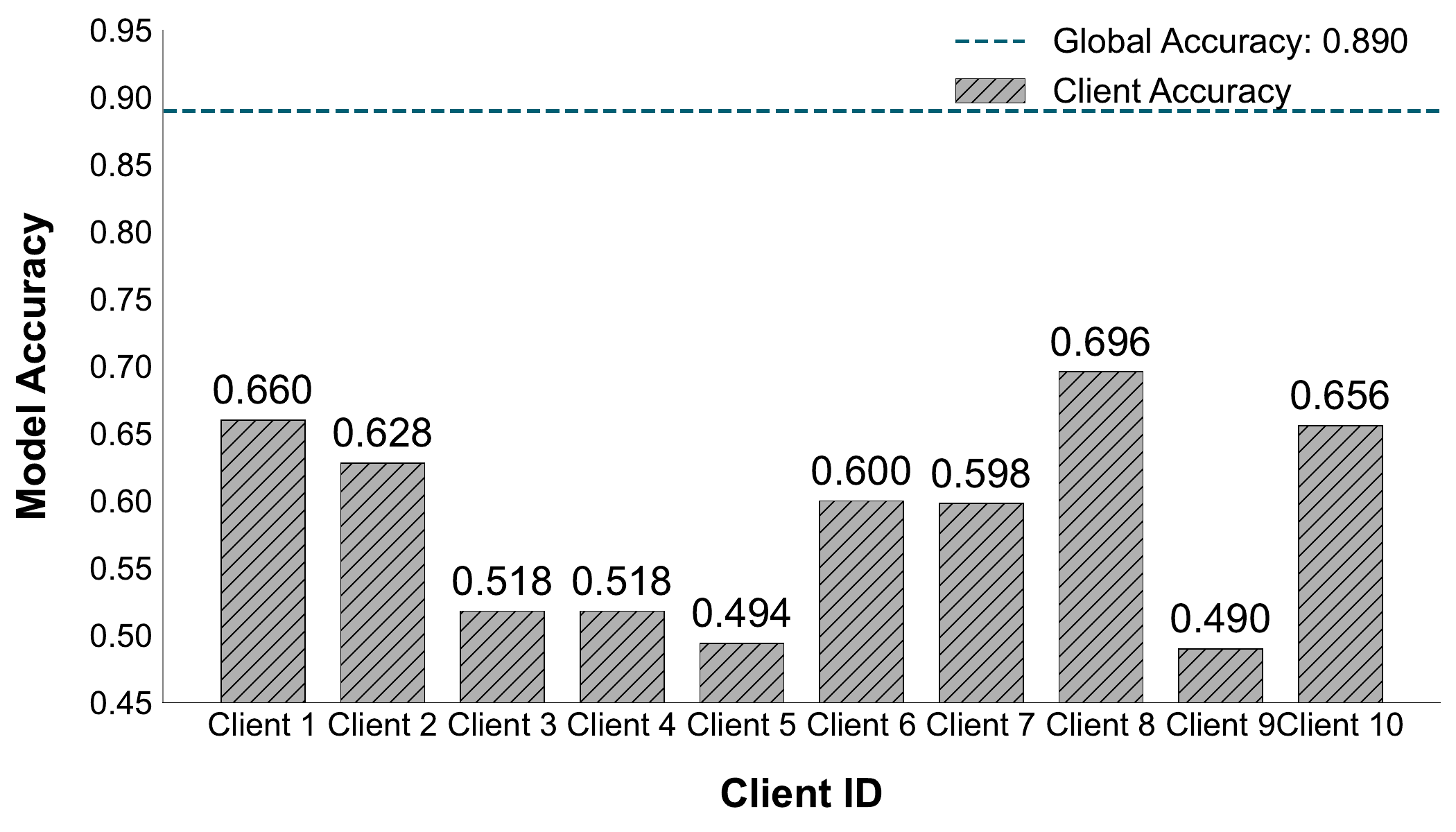}
  \caption{SUBJ}
\end{subfigure}

\caption{Performance comparison between local and global injection coefficients. The figure shows the accuracy of each client when using its locally optimized coefficient versus the globally aggregated coefficient obtained through federated averaging.}
\label{fig:bar-performanes}
\end{figure*}

\begin{figure*}[htbp]
  \centering
  \begin{subfigure}[b]{0.32\textwidth}
    \centering
    \includegraphics[width=\textwidth, height=4.4cm, keepaspectratio]{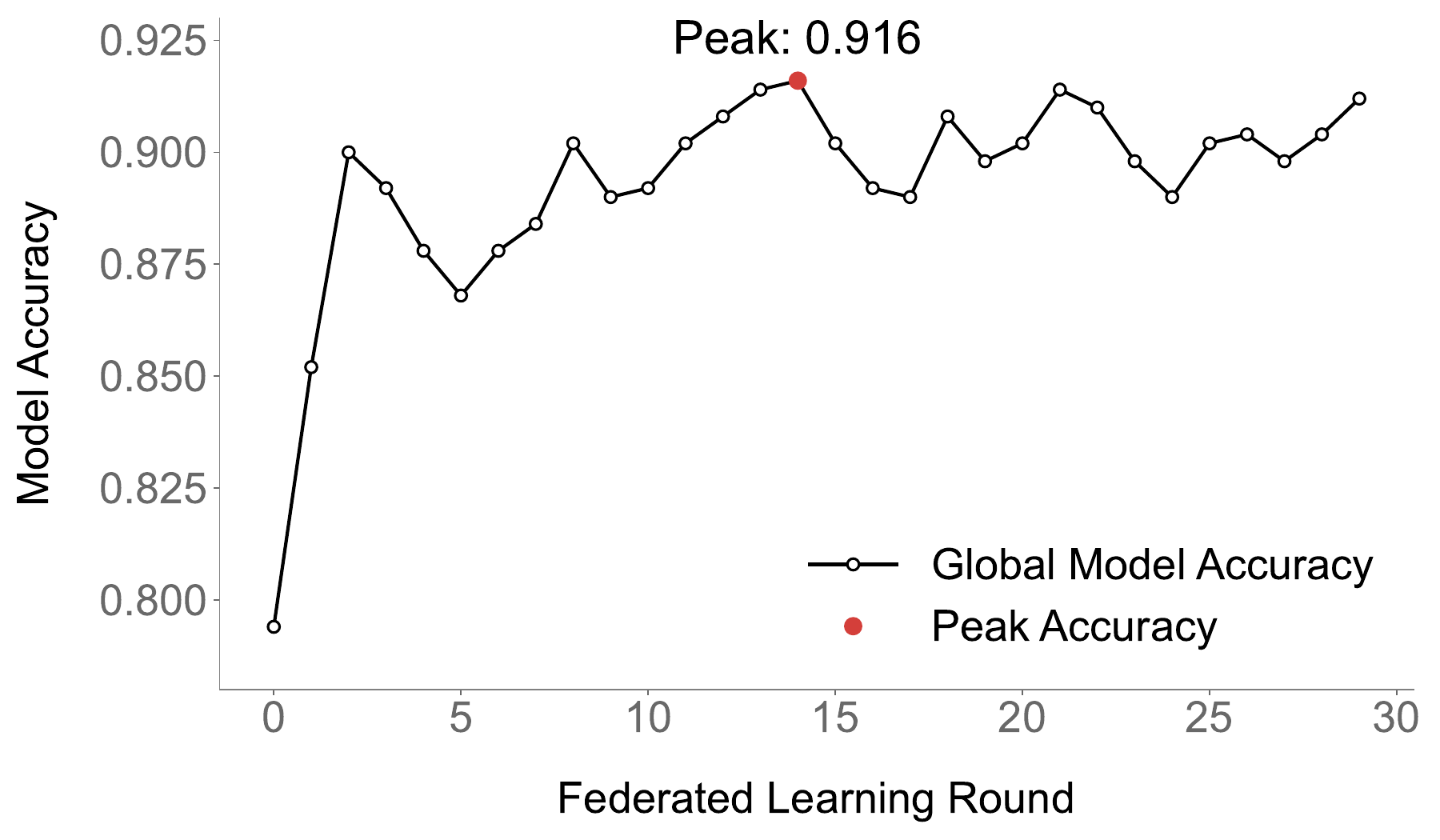}
    \caption{AG News}
  \end{subfigure}
  \hfill
  \begin{subfigure}[b]{0.32\textwidth}
    \centering
    \includegraphics[width=\textwidth, height=4.4cm, keepaspectratio]{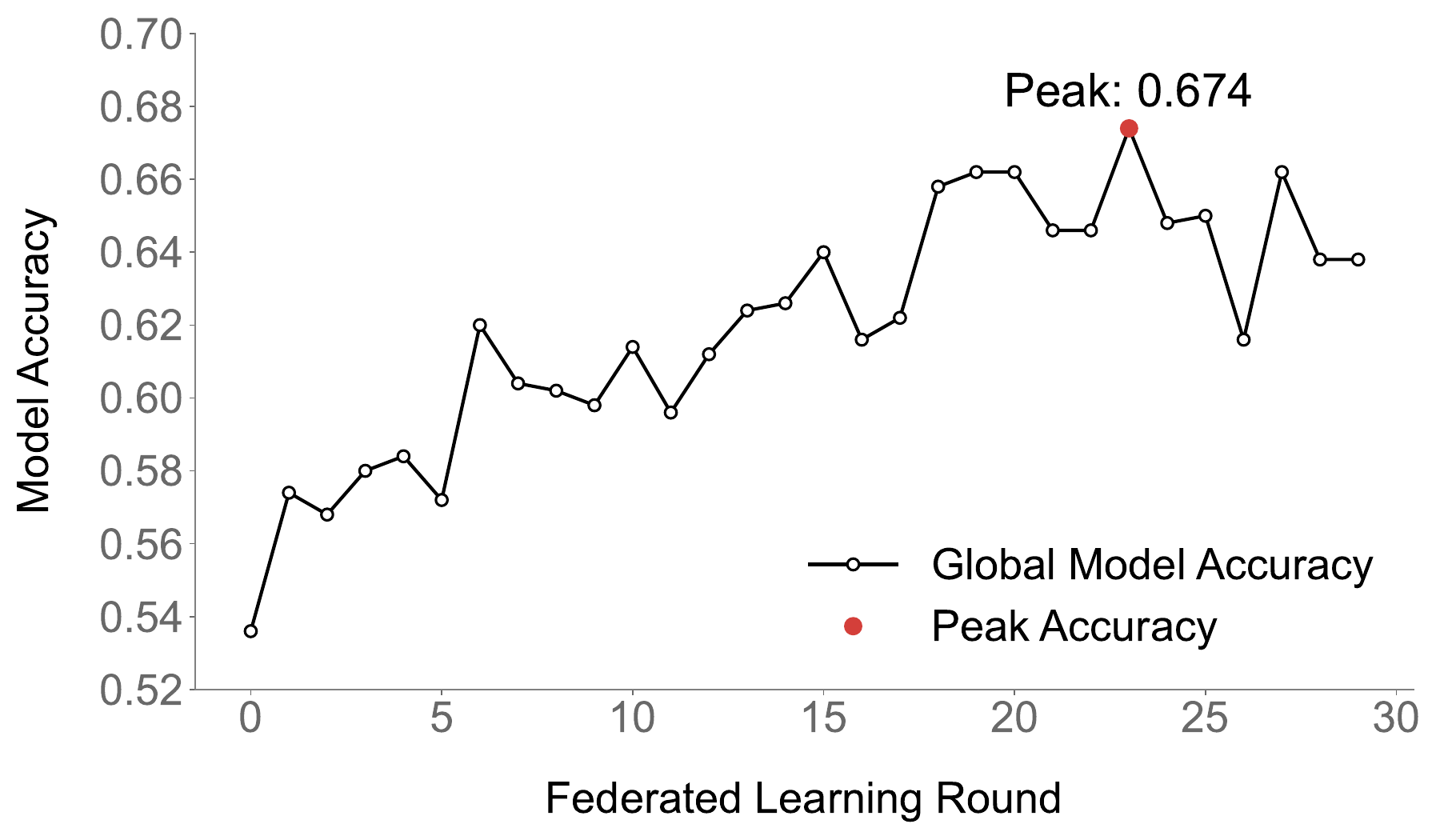}
    \caption{Emotion}
  \end{subfigure}
  \hfill
  \begin{subfigure}[b]{0.32\textwidth}
    \centering
    \includegraphics[width=\textwidth, height=4.4cm, keepaspectratio]{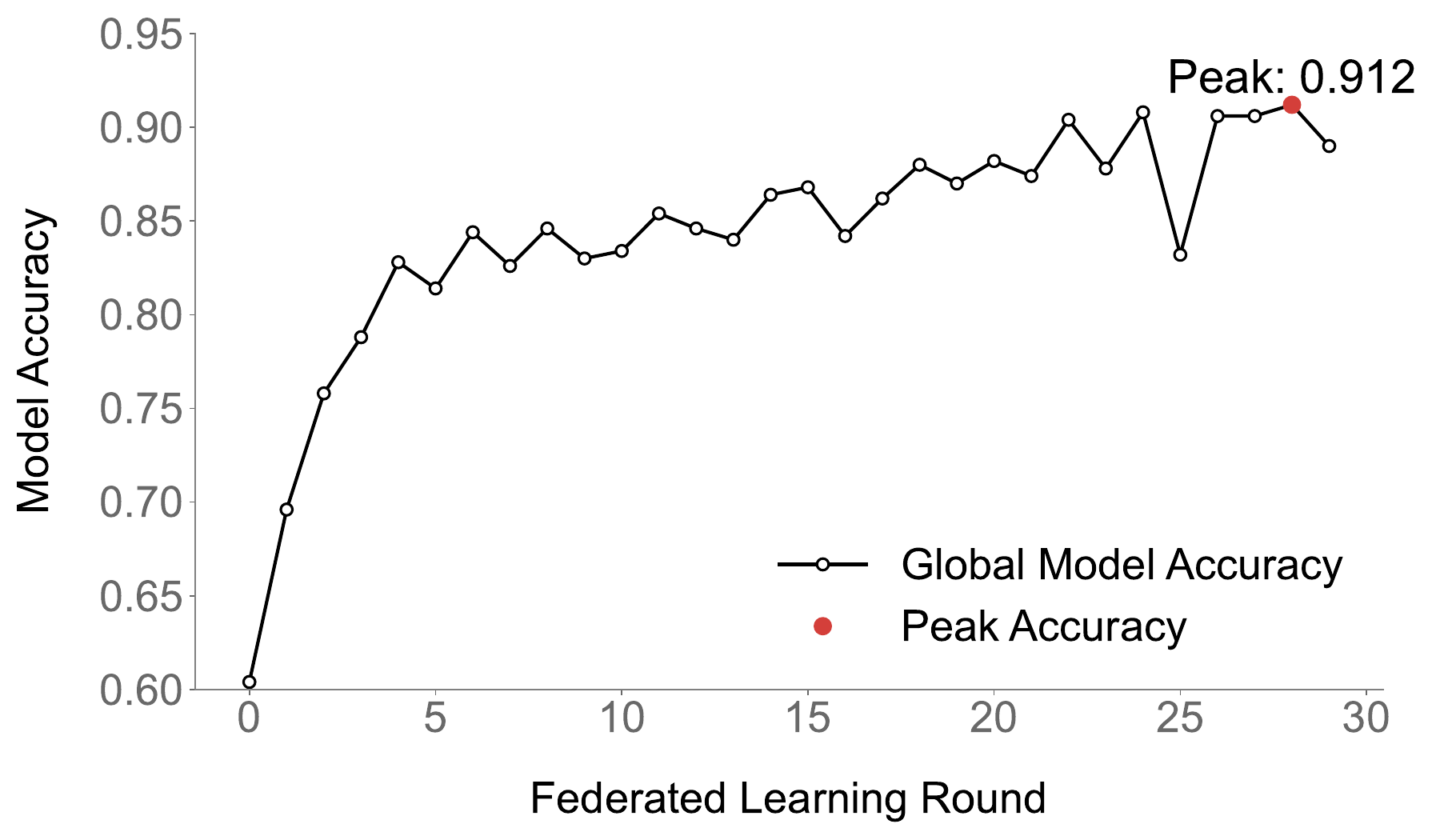}
    \caption{SUBJ}
  \end{subfigure}
  \caption{Analysis of the impact of injection coefficient optimization on performance. The figure illustrates how model performance evolves as the injection coefficient is optimized over successive communication rounds.}
  \label{fig:Tred-performanes}
\end{figure*}
\subsection{Efficiency and Communication Evaluation}

In terms of system efficiency, we compare the communication and clients computation overhead of IFed-ICL with the mainstream PEFT baseline FedAvg-LoRA. Communication overhead is defined as the total amount of data (in kilobytes, KB) each client uploads to the server per round. Clients computation overhead refers to the total time (in seconds) required to complete one local task per round.

Tables~\ref{tab:communication_breakdown_detailed} and~\ref{tab:runtime_only} present the efficiency comparison between IFed-ICL and FedAvg-LoRA from the perspectives of communication and computation. In terms of communication, IFed-ICL exhibits a decisive advantage. As detailed in Table~\ref{tab:communication_breakdown_detailed}, during the core training phase, IFed-ICL requires only 1.8 KB of communication per round to transmit a lightweight injection coefficient, while FedAvg-LoRA needs to exchange approximately 13.08 MB of LoRA weight matrices. Even accounting for the one-time transmission of context vectors during initialization (approximately 514 KB), our method remains highly efficient. 

From the perspective of computation, IFed-ICL further improves efficiency by restricting backpropagation to a minimal set of injection coefficients, thereby significantly reducing client-side complexity. As shown in Table~\ref{tab:runtime_only}, IFed-ICL achieves 20–30 times faster computation speeds on average, and up to 41.22 times in the best case compared to FedAvg-LoRA. While IFed-ICL takes slightly longer than Zero-Shot and Local ICL. Moreover, IFed-ICL provides the dual benefits of privacy preservation and performance enhancement under a federated learning setting, unlike Zero-Shot and Local ICL, which are more suitable for standalone or trusted environments and are difficult to deploy in real-world federated scenarios. Thus, the computational cost of IFed-ICL can be considered a necessary and acceptable trade-off in the privacy–performance balance.

The core innovation of IFed-ICL lies in exchanging only a small number of low-dimensional scalar coefficients, rather than full-scale model weights. This drastically reduces both communication and computation burdens. Such a property not only alleviates deployment bottlenecks in bandwidth-constrained or high-latency environments, but also provides practical communication feasibility for large-scale applications on mobile, edge, and IoT devices in real-world federated settings.

\subsection{Impact of Federated Aggregation on Injection Coefficient Performance}

To evaluate the effectiveness of forming a global injection coefficient through federated aggregation of locally calibrated coefficients, we compare the accuracy of the global coefficient on each client with that of locally optimized injection coefficients derived independently using only local data. For each round, we record the accuracy of all 10 client-specific models, their average accuracy, and the accuracy of the global model. By quantifying the difference between the global model accuracy and the mean local model accuracy, and analyzing the distribution of individual local model performance, we assess the impact of federated aggregation on overall model performance and generalization.

As illustrated in Figure~\ref{fig:bar-performanes}, the federated aggregation mechanism in IFed-ICL substantially improves the performance of the global injection coefficient. Compared to the average accuracy of local models, the global model achieves notable improvements of approximately 10.71\%, 26.05\%, and 12.81\% on the AG News, SUBJ, and Emotion datasets, respectively. Moreover, in most rounds, the global model outperforms the majority of individual local models. By integrating knowledge from heterogeneous clients, the aggregation process yields a superior global coefficient that significantly enhances model performance and generalization. This effect is particularly pronounced on the SUBJ dataset, which features highly non-uniform data distributions, thereby demonstrating the efficiency and robustness of the proposed framework in real-world federated learning scenarios.

As illustrated in Figure~\ref{fig:Tred-performanes}, we analyze the performance trajectory and stability of the global injection coefficient as the number of federated rounds increases. The experimental results indicate a consistent improvement in model performance over successive rounds, suggesting that the iterative refinement of the global injection coefficient plays a pivotal role in enhancing the effectiveness of IFed-ICL through collaborative optimization.

\section{Conclusion}

This paper proposes a Implicit Federated In-context Learning framework,  which decomposes the federated process into two components: context vector aggregation and injection coefficient optimization. This enables lightweight task adaptation for LLMs. Specifically, each client is responsible for transforming the context into vector representations and calibrating the injection coefficients through multi-round federated optimization based on local data. A one-time linear injection is then performed to achieve model adaptation. Unlike traditional approaches such as context concatenation or full model fine-tuning in federated settings, IFed-ICL decouples data from model training, significantly reducing both communication and computation costs. Experiments across multiple text classification tasks demonstrate the effectiveness of the proposed framework, offering a new paradigm for distributed intelligent collaboration on resource-constrained devices.

\bibliography{references}


\end{document}